\documentclass[sn-mathphys,iicol,Numbered]{sn-jnl}% Math and Physical Sciences Reference Style

\usepackage{graphicx}%
\usepackage{multirow,multicol}%
\usepackage{amsmath,amssymb,amsfonts}%
\usepackage{breqn}
\usepackage{amsthm}%
\usepackage{mathrsfs}%
\usepackage{verbatim}
\usepackage[title]{appendix}%
\usepackage{xcolor, soul}%
\soulregister\cite7 % 针对\cite命令
\soulregister\citep7 % 针对\citep命令
\soulregister\ref7 
\soulregister\etal7 
\usepackage{textcomp}%
\usepackage{manyfoot}%
\usepackage{booktabs}%
\usepackage{algorithm}%
\usepackage{algorithmicx}%
\usepackage{algpseudocode}%
\usepackage{import}
\usepackage{listings}%
\usepackage{wrapfig}
\newcommand{\etal}{\textit{et al.}}
\algnewcommand\INPUT{\item[\textbf{Input:}]}%
\algnewcommand\OUTPUT{\item[\textbf{Output:}]}

\begin{document}

\title[Article Title]{XnODR and XnIDR: Two Accurate and Fast Fully Connected Layers For Convolutional Neural Networks}

\author[1]{\fnm{Jian} \sur{Sun}} \email{Jian.Sun86@du.edu}

\author[2]{\fnm{Ali} \sur{Pourramezan Fard}} \email{Ali.Pourramezanfard@du.edu}

\author*[3]{\fnm{Mohammad} \sur{H. Mahoor}} \email{mohammad.mahoor@du.edu}


\affil[1]{\orgdiv{Department Of Computer Science}, \orgname{University of Denver}, \orgaddress{\street{2155 E Wesley Ave}, \city{Denver}, \postcode{80210}, \state{CO}, \country{United States of America}}}

\affil*[2,3]{\orgdiv{Department Of Computer Engineering}, \orgname{University of Denver}, \orgaddress{\street{2155 E Wesley Ave}, \city{Denver}, \postcode{80210}, \state{CO}, \country{United States of America}}}


\abstract{Capsule Network is powerful at defining the positional relationship between features in deep neural networks for visual recognition tasks, but it is computationally expensive and not suitable for running on mobile devices. The bottleneck is in the computational complexity of the Dynamic Routing mechanism used between the capsules. On the other hand, XNOR-Net is fast and computationally efficient, though it suffers from low accuracy due to information loss in the binarization process. To address the computational burdens of the Dynamic Routing mechanism, this paper proposes new Fully Connected (FC) layers by xnorizing the linear projection outside or inside the Dynamic Routing within the CapsFC layer. Specifically, our proposed FC layers have two versions, XnODR (Xnorize the Linear Projection Outside Dynamic Routing) and XnIDR (Xnorize the Linear Projection Inside Dynamic Routing). To test the generalization of both XnODR and XnIDR, we insert them into two different networks, MobileNetV2 and ResNet-50. Our experiments on three datasets, MNIST, CIFAR-10, and MultiMNIST validate their effectiveness. The results demonstrate that both XnODR and XnIDR help networks to have high accuracy with lower FLOPs and fewer parameters (e.g., 96.14\% correctness with 2.99M parameters and 311.74M FLOPs on CIFAR-10).}

\keywords{CapsNet; XNOR-Net; Dynamic Routing; Binarization; Xnorization; Machine Learning; Neural Network}

\maketitle

\section{Introduction} \label{sec:1}

% presents the problem and research target 
The trade-off between model accuracy and processing speed is an interesting research topic in the field of Machine Learning (ML). CapsuleNet (CapsNet)~\cite{A1_caps} is a neural network architecture that embeds Dynamic Routing into Convolutional Neural Networks (CNNs). XNOR-Net~\cite{A2_xnor} is a neural network that utilizes Xnorization in the CNNs. CapsNet~\citep{A1_caps} pays more attention to accuracy and less concerns about speed, while XNOR-Net~\citep{A2_xnor} aims to improve computational speed and maintain moderate performance. 

This paper aims to investigate and implement a novel deep neural network architecture that merges the benefits of CapsNet and XNOR-Net. It presents the opportunity to attain high accuracy, albeit with a trade-off in terms of processing speed.

% Briefly introduce CapsNet
CapsNet~\citep{A1_caps} was designed to address the shortcoming of CNNs, which are weak at capturing global information, especially the positional relation between features in images. For example, a face with randomly ordered eyes, ears, nose, mouth, and eyebrows will be wrongly recognized as a human face by CNN model~\cite{A62_face_align}. A bigger kernel size can be the solution, but it causes a surge in computational costs. Later, Sabour~\etal~\cite{A1_caps} developed dynamic routing to update capsules so that CapsNet studies the positional relation between features. A capsule is a group of neurons whose activity vector represents the instantiation parameters of a specific type of entity such as an object or an object part~\cite{A1_caps}. In better words, a capsule is a vector, where its length size means the possibility of the appearance of an object or an image property. Its direction represents the object's image property, such as location, shape, size, direction, etc. The Dynamic Routing (DR) mechanism, derived from K-Means clustering, recursively computes and updates capsules. Sabour~\etal~embedded DR into the classifier and named this classifier the CapsFC layer.

% Briefly introduce XNOR-Net
On the other hand, XNOR-Net~\cite{A2_xnor} was designed to use binary operations in the CNN layers of a network using the xnorization operation~\cite{A2_xnor}. Xnorization binarizes both input tensor and weight into the sign vectors and the scale factor first, then it calculates binary dot product with the XNOR operator rather than multiplication between two sign vectors and times those two scale factors. Xnorization applies only addition instead of multiplication in the Conv layer to reduce computational times and fasten the calculation (Section~\ref{sec:3.3} and Appexdix~\ref{appendix:A}). It is a simple and efficient approximation to Conv operation~\citep{A2_xnor}. Rastegari~\etal~\citep{A2_xnor} named the xnorized layer the XNORConv layer.

% CapsFC structure and Problem of CapsNet, and mention the part to modify
However, CapsNet and XNOR-Net are far from impeccable and have their own limitations. CapsFC layer contains a linear projection and DR (Section~\ref{sec:3.1}). Sabour~\etal~expand the 3-dimensional input tensor into a 5-dimensional one before linear transformation, which increases parameters, multiplication, and addition operations (MADD~\cite{A18_MBV2}), and processing time. Simultaneously, DR also has this problem. Its routing iteration is a looping process, which multiplies the computational times (Section~\ref{sec:3.1}). Thus, compared with the usual Fully Connected (FC) layers, the CapsFC layer is slow on both training and inference due to the dimension expansion and routing iteration. For example, Tables~\ref{tab:all-acc} and~\ref{tab:all-flops-para} show that CapsNet has 99.65\% accuracy with 6.80M parameters on MNIST, while MobileNetV2 achieves 99.32\% accuracy (a comparable result) with a network containing 3.05M parameters (less than half of CapsNet's parameter). Furthermore, the large number of network parameters and its slow inference speed weakens the effect of CapsNet on mobile devices.

\begin{figure*}[t]
\begin{center}
\includegraphics[width=0.8\linewidth,height=0.17\textheight]{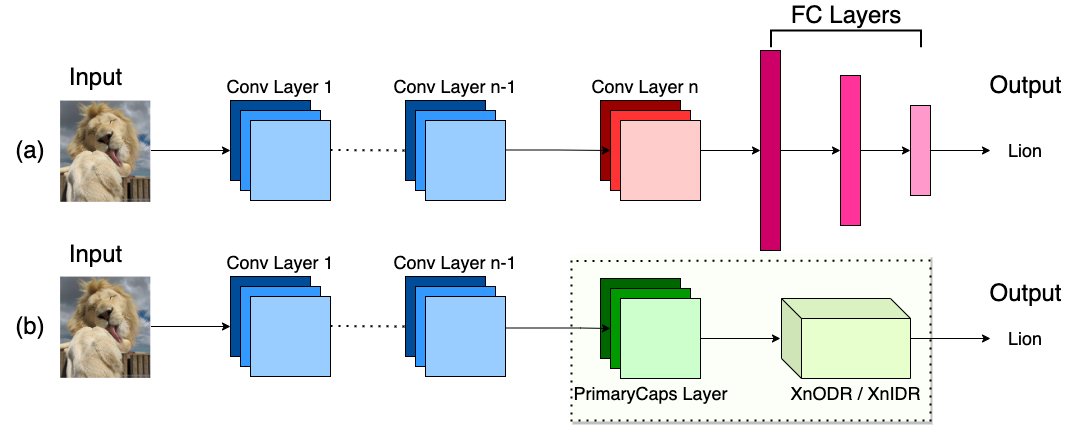}
\end{center}
   \caption{(a) is the structure of typical CNN-based models; (b) is the structure of models with XnODR/XnIDR. The layers within the light green box are the modified part. We exchange the last convolutional layer and all FC layers with PrimaryCaps layer and XnODR/XnIDR.}
\label{fig:m-struct}
\end{figure*}

% XNOR-Net structure and Problem of XNOR-Net, and mention the part to modify

On the other hand, XNOR-Net achieves reasonable accuracy with fewer operations and faster speed on MNIST~\cite{A2_xnor}. It is also an undeniable truth that XNOR-Net has information loss due to Xnorization that results in lower accuracy compared to the full-precision AlexNet on ImageNet~\cite{A2_xnor, A48_img_net}. Usually, ImageNet requires a deep neural network, which has more convolution layers, such as CoCa and CoAtNet-7~\cite{A73_Coca, A72_CoAtNet}. However, the more layers XNOR-Net xnorizes, the more information it loses. Too much information loss impairs the XNOR-Net to classify large-size images (like 128$\times$128 and 224$\times$224), small object images, or complex images~\cite{A2_xnor}.

% Start fusing both models and present experimental results
By digging deep into CapsNet and XNOR-Net, we learn that the CapsFC layer and Xnorization are the core of their success. A more stable module may appear when applies Xnorization in the CapsFC layer. The followings are more specific discussions.

Sections~\ref{sec:3.1} and~\ref{sec:3.3} show that there are two linear projections, $LP_{out}$ and $LP_{in}$, in the CapsFC layer. $LP_{out}$ is outside the DR, while $LP_{in}$ is inside it. Either one involves heavy computation. Decreasing either one's computational times will speed up the model. Then, the calculation in the FC layer is the linear projection, which equals to Conv layer with the $1\times 1$ kernel. the aforementioned Xnorization is to simplify and approximate the Conv layer. Therefore, we decided to deploy Xnorization on the $LP_{out}$ and $LP_{in}$ from the CapsFC layer separately and propose new FC layers. Specifically, we call the FC layer XnODR if it \textbf{xn}orizes the linear projection \textbf{o}utside \textbf{D}ynamic \textbf{R}outing. Otherwise, we call it XnIDR if it \textbf{xn}orizes the linear projection \textbf{i}nside \textbf{D}ynamic \textbf{R}outing. We envision XnODR and XnIDR that enables the network to maintain comparable or higher accuracy like CapsNet while reducing the parameters and floating point of operations (FLOPs) and increasing the network speed like XNOR-Net.

To test the generalization of XnODR and XnIDR, we utilize XnODR/XnIDR separately to replace the usual FC layers of the typical lightweight model, MobileNetV2. In addition, we do the same procedure on the heavyweight ResNet-50 model. Fig.~\ref{fig:m-struct} shows the way to embed XnODR/XnIDR into CNN-based models. We validate these variants on MNIST, CIFAR-10, and MultiMNIST datasets. Section~\ref{sec:4} shows that XnODR and XnIDR can properly replace the dense layers on the lightweight and heavyweight models. Moreover, both XnODR and XnIDR help speed up the model computation and maintain comparable or even better accuracy.

% list contributions and route map
Overall, the contributions of this paper are as follows:

\begin{itemize}
\item We propose a new Fully Connected Layer called XnODR by deploying Xnorization on the CapsFC layer. To be specific, we xnorize the linear projection outside the dynamic routing.
\item We propose a new Fully Connected Layer called XnIDR by xnorizing another linear projection inside the dynamic routing.
\item XnODR and XnIDR improve the performance (i.e., better accuracy, less FLOPS, and parameters) of both lightweight (MobileNetV2) and heavyweight (ResNet-50) models.
\end{itemize}

The remainder of the paper is organized as follows. Section~\ref{sec:2} provides an overview of related work. Section~\ref{sec:3} explains the new proposed FC layers. Section~\ref{sec:4} introduces databases used in this work and presents the experimental configuration, evaluation metrics, experimental results, ablation study, and analyses. We discuss our work in Section~\ref{sec:5} and finally conclude the paper in Section~\ref{sec:6}.

\section{Related Work} \label{sec:2}
\subsection{Capsule Network} \label{sec:2.1}

\begin{figure} [ht]
\centering
\caption{CapsNet Structure. $\hat{I}_{\text{Cap}}$ is the predicted capsules, $I_{\text{Cap}}$ is the updated Capsules, $b$ represents all temporary values, and $v$ represents the activated capsules. Then, $c$ is the coupling coefficient tensor, $c_{ij}$ measures the probability that $\hat{I}_{\text{Cap}_{i}}$ activates $\hat{I}_{\text{Cap}_{j}}$. $N$ is the iteration number.}
\vspace{0.2cm}
\includegraphics[width=6.5cm,height=8cm]{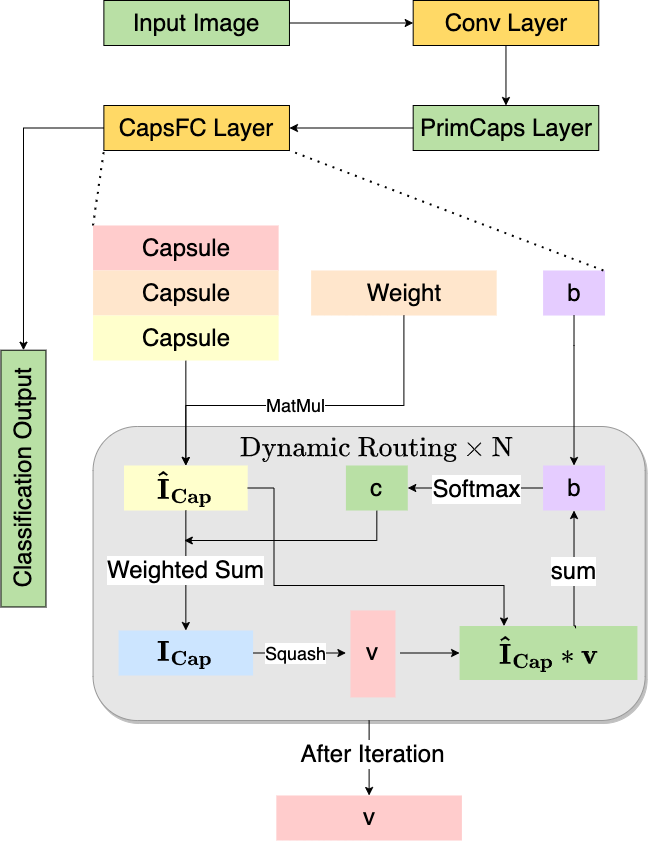}
\label{fig:capsnet}
\vspace{-0.3cm}
\end{figure}

CNNs focus more on extracting features from images rather than orientational and relative spatial relationships between those features. The max-pooling layer helps CNNs to work surprisingly better than previous CNN models. The max-pooling layer increases the CNNs' performance because it suppresses the noise and ignores less significant elements. However, this property also makes CNNs lose much valuable information, especially positional features such as where things are. The contribution of the max-pooling layer, thus, covers CNNs' orientational and relative spatial relationships problem. CapsNet has solved this problem~\cite{A1_caps}. It consists of Convolutional layers, a PrimaryCapsule layer, a CapsFC layer, and a decoder~\footnote{This work focuses on the change of accuracy and speed of networks instead of the Decoder part}. The PrimaryCaps layer consists of the convolutional operation and max-pooling. The effect of the CapsFC layer is a classifier. Different from the typical FC layer, the CapsFC layer makes up of a linear projection (an affine transformation) and the DR mechanism (iteratively weighted-sum), which significantly improves the accuracy and interpretation of classification. Fig.~\ref{fig:capsnet} shows CapsNet's structure. In better words, the CapsFC layer is the core of CapsNet.

Researchers, who were inspired by CapsNet, improved the original CapsNet with different ideas, such as better performance on complex datasets~\citep{A20_variant_squash}, providing solid equivariance and invariance property~\citep{A21_grp_caps}, and detecting objects from features on a one-dimensional linear subspace~\citep{A22_sp_caps}. Then, Aff-CapsNets improved the robustness of affine transformations and dropped the dynamic routing mechanism~\citep{A24_Rob_Caps}. MRCapsNet and Res-CapsNet focused on enhancing feature extraction capability~\citep{A53_MRCapsNet, A54_Res-CapsNet}. The aforementioned papers pay more attention to elevating the accuracy of the CapsNet.

Other researchers have explored the other possibilities of CapsuleNet, such as addressing the visual distortion problem~\citep{A44_CapsnetSIFT}, fine-grained classification~\citep{A45_fine_grain}, text classification~\citep{A46_textcls}, wind speed prediction~\citep{A55_MOHHO}, and environmental monitoring~\citep{A56_PM25}. These papers extended the good performance of CapsNet to other tasks beyond purely image classification.

In our work, we focus on improving the speed of the CapsFC layer. With CapsFC layer, the CapsNet is time-consuming to train and run inference, especially on complex datasets like CIFAR-10, because of the DR's iterative structure and large-scale floating-point operations during convolutional calculation and linear projection. This paper presents a solution to address the low-speed performance of the CapsFC layer.

\subsection{XNOR Network} \label{sec:2.2}

The large number of parameters in a CNN usually causes inefficient computation and memory over-utilization. Researchers have proposed several methods to address this problem.

Some researchers proposed the theory of Shallow networks and did related experiments~\cite{A29_Cybenko, A30_Seide, A31_Dauphin, A32_Ba}. The core idea of Shallow networks is to mimic deep neural networks to get similar numbers of parameters and equivalent accuracy. Otherwise, Shallow networks return less comparable accuracy on ImageNet~\cite{A31_Dauphin}.

It is also sensible to assemble CNNs with compact blocks that cost less memory and have less FLOPs. For example, GoogleNet, ResNet, and SqueezeNet proposed new layers or structures and achieved several benchmarks with the cost of fewer parameters~\cite{A33_GoogleNet, A19_RES50, A34_SquNet}. Then, HGCNet was capable to elevate representation capability by fusing feature maps from different groups~\cite{A43_high_eff}.

Since CNNs can achieve good performance without the need for high-precision parameters, parameter quantization is a viable option to speed up the network computation. Therefore, researchers proposed many novel ideas, such as quantizing the weights of FC layers~\citep{A35_Vector_Q}, using ternary weights and 3-bits activations~\citep{A36_fix}, only quantizing neurons during the back-propagation process~\citep{A37_mul}, and vector quantization method~\citep{A47_vec_quant}.

Other researchers focused on improving either the accuracy or the speed of the networks using the quantization methods as well~\cite{A39_grdy, A40_low_bit, A41_prune}.

\begin{figure} [ht]
\centering
\caption{Xnorization process. $X$ and $W$ are input tensor and weight variables. $B_{X}$ and $B_{W}$ are binarized sign vectors, while $\alpha_{X}$ and $\alpha_{W}$ are scale factors.}
\includegraphics[width=5.5cm,height=5.cm]{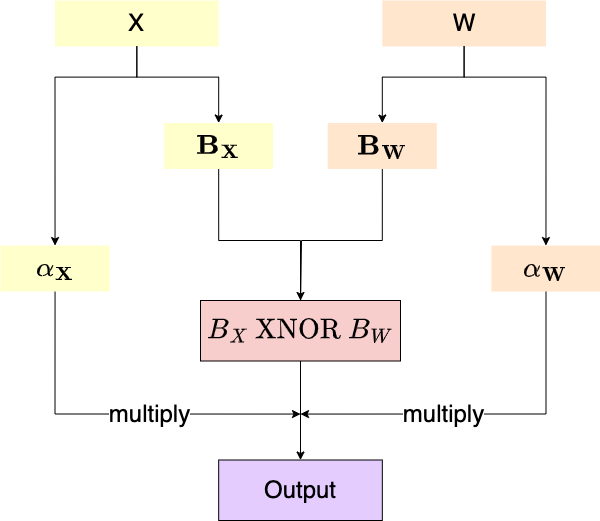}
\vspace{-0.3cm}
\label{fig:xnorization}
\end{figure}

XNOR-Net uses standard deep architectures instead of shallow ones and trains networks from scratch rather than implementing pre-trained networks or networks with compact layers. Moreover, it quantizes the weights and input values with two factors, +1, -1, instead of +1, 0, -1~\citep{A38_probnd}. \cite{A2_xnor} stated that the typical CNNs would cost more time as the size of tensors increases because that causes more multiplication and division operations while doing the convolutional calculations. To reduce the processing time and maintain the prediction accuracy, \cite{A2_xnor} proposed a new concept called Xnorization (Appendices~\ref{appendix:A} and~\ref{appendix:B}). Fig.~\ref{fig:xnorization} shows its structure. The advantage of the XNOR operation is that it uses plus and minus to do convolutional calculations rather than multiplication and division. Therefore, XNOR can save substantial processing time during training time. It is worth mentioning that XNOR-Net has comparable performance on MNIST and CIFAR10 compared to BNN(Binary Neural Network)~\citep{A2_xnor}. Moreover, XNOR-Net is more accurate than Binary Weight Network (BWN) on ImageNet~\citep{A48_img_net}.

XNOR-Net has many variants too. Ternary Sparse XNOR-Net~\cite{A25_ternary_xnor}, XNOR-Net++~\cite{A26_x++}, and Bi-Real-Net are good examples~\citep{A27_Bi_Real}. These models put effort into improving model representational capability. Zhu~\etal~proposed XOR-Net to offer a pipeline for binary networks both with and without scaling factors~\citep{A28_XOR}. The above papers aimed to reduce memory usage, speed up the inference time, and improve accuracy by creating different quantization methods or the new pipeline. Simultaneously, XNOR-Net has broad prospects for application as well, such as bird sound detection~\citep{A57_Bird_Sound_Detection}, reducing the impact of RRAM noise, and improving energy efficiency~\citep{A58_RRAM}.

We also found that Xnorization causes the network to lose much information. Insufficient information prevents XNOR-Net from achieving as high accuracy as CNNs, such as MobileNetV2 and ResNet-50. In this work, we focus on improving accuracy. Our idea is to add a routing mechanism rather than modifying the sign function, scaling factor, or the CNNs' main body. In this paper, CNNs' main body represents the CNNs without any FC layer.

\subsection{Different Routing Mechanisms} \label{sec:2.3}

Designing a new routing mechanism becomes popular after the appearance of DR. EM Routing algorithm~\citep{A3_emr} and VB Routing method~\citep{A23_vb} are good examples. EM Routing is more accurate than DR with the help of the Expectation-Maximization algorithm and Gaussian Mixture model, but it becomes more time-consuming than DR because of expanding the capsule to a 2-dimensional pose matrix and more for loops. People without many computing resources may hesitate to select EM Routing to train large-scale datasets such as ImageNet. VB Routing, on the other hand, has more flexible control over capsule complexity by tuning priors to induce sparsity and reducing the variance-collapse singularities inherent to MLE-based mixture models such as EM. Generally, VB Routing helps overcome overfitting and unstable training issues found in EM Routing. Our work mostly focuses on speeding up the routing process rather than developing a more crafted routing mechanism. On the other side, if our idea works, we would like to expand it to other routing mechanisms.

\section{Two New Fully Connected Layers: XnODR and XnIDR} \label{sec:3}

This section introduces the concept of XnODR and XnIDR and the steps to fuse the CapsFC layer and the Xnorization. Both XnODR and XnIDR obtain the properties of high accuracy from CapsNet and fast inference speed from XNOR-Net, and they are modified based on CapsFCLayer. See Section~\ref{sec:3.3} for more details.

\subsection{CapsFC Layer Review} \label{sec:3.1}

CapsFC layer has two parts, affine transformation and DR. DR~\citep{A1_caps} helps to activate core capsules, suppresses unimportant ones from lower layers, and highlights the core capsules with high probability. At the same time, as an iterative process, it improves performance by iteratively updating the output capsules. It is used in XnODR and XnIDR as the routing mechanism too. Next, we describe its concept in detail.

Traditional neurons have three steps, weighting, sum, and nonlinearity activation, which can be summarized as:

\begin{equation}\small \label{eq:usual-fc}
\begin{split}
a_{j} &= \sum_{i} w_{i}x_{i}+b \\
h_{j} &= f(a_{j})
\end{split}
\end{equation}
where $x_{i}$ is the input value from the neuron $i$, $w_{i}$ is a randomly initialized but trainable weight $i$ for $x_{i}$, $b$ is the bias, $a_{j}$ is the $j^{th}$ neuron's output of linear projection, $f$ is the nonlinearity activation function, $h_{j}$ is the $j^{th}$ neuron's final output.

In the CapsFC layer, \cite{A1_caps} took the capsules as the input variable and designed a new activation function, the Squash function. Then, Sabour~\etal~added affine transformation before the weighting step. This makes the CapsFC layer have four steps, affine transformation, weighting, sum, and nonlinearity activation, which can be summarized as:

\vspace{-0.5cm}
\begin{align}\tiny \label{eq:aff-trans}
\hat{C}_{out_{j|i}} &= W_{ij}C_{out_{i}} \\
\begin{split} \label{eq:wght-sum}
C_{out_{j}} &= \sum_{i}c_{ij}\hat{C}_{out_{j|i}} 
\end{split}\\ \label{eq:squash}
v_{j} &= \frac{||C_{out_{j}}||^{2}}{1+||C_{out_{j}}||^{2}}\cdot \frac{C_{out_{j}}}{||C_{out_{j}}||}
\end{align}
where $C_{out_{i}}$ denotes the capsule $i$ from lower layer $l$, and $W_{ij}$ denotes the weight matrix between capsule $i$ and $j$. $\hat{C}_{out_{j|i}}$ is the prediction capsules from layer $l$ to layer $l+1$. Here, Eq.~\ref{eq:aff-trans}, denoted as $LP_{out}$, is the affine transformation. The vectors $C_{out_{i}}$ are multiplied by the corresponding weight matrices $W_{ij}$ that encode important spatial and other relationships between the lower-level features (eyes, mouth, and nose) and higher-level feature (face). The meaning of affine transformation here is to observe the object from different views and angles. Then, multipling $C_{out_{i}}$ by $W_{ij}$ returns the predicted capsules $\hat{C}_{out_{j|i}}$.

Next, in Eq.~\ref{eq:wght-sum}, $c_{ij}$ represents the weight that multiplies the predicted vector $\hat{C}_{out_{j|i}}$ from the lower-level capsules and serves as the input to a higher level capsule. In simple terms, $c_{ij}$ is the coupling coefficient that depicts the relationship between capsule $i$ and capsule $j$. $C_{out_{j}}$ is the output of capsule $j$ from layer $l+1$. In the meanwhile, $c_{ij}$ measures the probability that $C_{out_{i}}$ activates $C_{out_{j}}$. The parameter $c_{ij}$ is determined by the Softmax function and a new coefficient $b_{ij}$ (Eq.~\ref{eq:cpl-cffcnt}), which is a temporary value that is updated iteratively. The sum of $c_{ij}$ over $j$ is 1. At the start of the training, the value of $b_{ij}$ is initialized to zero.

\begin{equation} \label{eq:cpl-cffcnt}
c_{ij}\ =\ \frac{exp(b_{ij})}{\sum_{k}exp(b_{ik})}
\end{equation}

Finally, Sabour~\etal~apply the \textbf{Squash} function (Eq.~\ref{eq:squash}), an activation function like ReLU, to activate the output capsules. $||C_{out_{j}}||$ is the L2-norm of $C_{out_{j}}$. The parameter $v_{j}$ is the output of capsule $j$ from layer $l+1$ after squashing. The Squash function controls the size of $C_{out_{j}}$ to be less than 1 and preserves its direction. $v_{j}$ is the final output vector of the capsule $j$, which represents the activated capsule. Here, $v_{j}$ also helps update $b_{ij}$, which can be formulated as:

\begin{equation}\small \label{eq:tmprry-val}
b_{ij}\ =\ b_{ij}\ +\ \hat{C}_{out_{j|i}}\cdot v_{j}.
\end{equation}

Eq.~\ref{eq:tmprry-val} shows that the new $b_{ij}$ equals the old $b_{ij}$ plus the dot product of the activated capsule $v_{j}$ and the predicted capsule $\hat{C}_{out_{j|i}}$. We denote $\hat{C}_{out_{j|i}}\cdot v_{j}$ as $LP_{in}$. The dot product looks at the similarity between predicted capsules and activated capsules. Also, the lower-level capsule will send its output to the higher-level one, whose output is similar to the predicted one. This similarity is captured by the dot product. The larger the dot product, the higher the correlation between the activated capsule and the predicted capsule is, and the larger the $b_{ij}$ is. Then, referring to Section~\ref{sec:3.3}, we modified Eq.~\ref{eq:aff-trans} in the proposed XnODR, while changing Eq.~\ref{eq:tmprry-val} to get XnIDR.

The aforementioned parameters get updated iteratively. According to \cite{A1_caps}, the model would perform well when setting the iteration number to 3.

In general, Sabour~\etal~summarized the above formulations to an iterative process called DR~\citep{A1_caps}. The iterative routing process implements the property of local feature maps to calculate and decide whether or not to activate capsules. Moreover, with the help of the capsules, DR takes the feature maps' location, direction, size, and other detailed information into consideration rather than simply detecting features such as CNNs. For example, we can make either a house or a sailboat with one square and one triangle. If we train the network by the house and test it on a sailboat, CNNs would wrongly classify it as a "house" since it only detects features independently~\cite{A65_Boat_House}. Oppositely, CapsNet with Dynamic Routing would activate related sailboat capsules, avoid mistakes after comprehensive analysis, and help to improve the prediction result by updating capsules in the FC layer. Section~\ref{sec:3.3} provides the defection analysis of DR.

\subsection{XNOR-Net Review} \label{sec:3.2}
 
XNOR-Net consists of a stack of XnorConvLayer. XnorConvLayer is an xnorized Conv layer, which binarizes input tensor and weight first and does the binary dot product with XNOR-Bitcounting operations. Binarization is to split the tensor into two parts. One is a sign matrix (spanned from 2 values $\{-1, 1\}$), and the other one is the scaling factor. This process, named Xnorization~\cite{A2_xnor}, is to approximate the common Convolutional operation. Thereby, \textbf{Binarization} and \textbf{XnorConvLayer} are two vital concepts for XNOR-Net~\cite{A2_xnor}. We also use them to define XnODR/XnIDR. Appendixes~\ref{appendix:A}~and~\ref{appendix:B} present more details on Binarization and XnorConvLayer. Next, we directly introduce XnODR and XnIDR. 

\subsection{XnODR and XnIDR} \label{sec:3.3}

Referring to Section~\ref{sec:1}, a usual dense layer only does one linear projection, while the CapsFC layer has $LP_{out}$ and $LP_{in}$. The usual dense layer only takes a 2-dimensional tensor as input, while $LP_{out}$ requires expanding the input feature from three dimensions to five dimensions before computation. Therefore, $LP_{out}$ costs more parameters, MADD~\cite{A18_MBV2}, and processing time than the original dense layer~\cite{A18_MBV2}. This expansion causes a time-consuming issue. Simultaneously, the $LP_{in}$ is in DR. The iterative structure of the DR mechanism increases MADD and consumes more time on inference, which slows down the model speed. Simply stated, to seek implicit information, CapsNet trades off accuracy with speed.

Given that "fully connected layers can be implemented by convolution"~\citep{A2_xnor}, which means that the FC layer is like a convolution layer with kernel size $1\times 1$~\citep{A2_xnor}, it is also feasible to binarize the FC layer in XNOR-Net. Binarization is a very convenient function. However, it averages the pixel values among each channel as a scaling factor that breaks the hierarchy of the pixel values and fails to collect many implicit features. Furthermore, it only approximates the pixels simply by the product of the sign matrix and scaling factor, which exacerbates the information loss, a very apparent negative influence. Thereby, Xnorization at different layers aggravates the network's information loss, which prevents XNOR-Net from performing as well as full-precision CNN-based models. This information loss slightly affects the XNOR-Net's performance on small-scale and simple datasets but heavily prevents XNOR-Net from detecting well on large-scale complex datasets such as ImageNet and AffectNet~\citep{A52_affnet}. For example, in ImageNet, compared to 56.6\% Top-1 accuracy at the full-precision AlexNet, AlexNet with Xnorization only achieves 44.2\% Top-1 accuracy. It is also very important to xnorize the correct layers. If we only xnorize the last Conv layer or the final dense layer, we may get as good accuracy as a full-precision model does. However, the accuracy is far from satisfying when xnorizing all the layers, especially shallow Conv layers like the second one. The purpose is to avoid too much information loss. The network already extracts enough feature maps before the last Conv layer or the final Dense layer but exactly starts mining features at the second layer. The model will lose more key features if it xnorizes the shallow Conv layers. Hence, it is more reasonable to implement Xnorization on one of the top layers instead of blindly deploying it on every layer to pursue speed.

\begin{table}[ht]\small
\centering
\caption{shows the size of variables for XnODR and XnIDR. $bs$ is the batch size of the input value, $\text{caps}_{in}$ is the number of capsules loaded into this layer, $\text{caps}_{out}$ is the number of the capsules output from this layer, $1$ means the capsule is a 1-dimensional vector, $\text{dim}_{in}$ is the element number of this each input capsule, $\text{caps}_{out}$ is the dimension of each output capsule.}
\begin{tabular}{l|c} 
\hline
Variables & Size \\
\hline
\hline
$I_{\text{Prim}}$ & [$bs$, $\text{caps}_{in}$, $\text{dim}_{in}$] \\
$I_{\text{Cap}}$ & [$bs$, $\text{caps}_{in}$, $\text{caps}_{out}$, $1$, $\text{dim}_{in}$]  \\
$W_{\text{Cap}}$ & [$\text{caps}_{in}$, $\text{caps}_{out}$, $\text{dim}_{in}$, $\text{dim}_{out}$] \\
$\hat{I}_{\text{Cap}}$ & [$bs$, $\text{caps}_{in}$, $\text{caps}_{out}$, $1$, $\text{dim}_{out}$] \\
$v$       & [$bs$, $1$, $\text{caps}_{out}$, $1$, $\text{dim}_{out}$] \\
$b$       & [$bs$, $\text{caps}_{in}$, $\text{caps}_{out}$, $1$, $1$] \\
\hline
\end{tabular}
\vspace{-0.3cm}
\label{tab:4}
\end{table}

Intuitively, we fuse the CapsFC layer from CapsuleNet and Xnorization from XNOR-Net to create a new FC layer, a more accurate and faster layer. During the training and inference, this layer implements Xnorization to simplify operations and speed up the model. It also helps maintain comparable prediction accuracy by taking advantage of Capsules and DR to extract the direction, location, and other sophisticated information among feature maps. Furthermore, the new FC layer would do binarization before the linear projector, and replace multiplications with additions and subtractions. In detail, we xnorize $LP_{out}$ to get the first new FC layer while xnorizing $LP_{in}$ to get the second one. In total, there are two versions.

Given that both Eqs.~\ref{eq:aff-trans} and~\ref{eq:tmprry-val} take one capsule as the input, we rewrite two equations as Eqs.~\ref{eq:n-aff-trans} and~\ref{eq:n-tmprry-val} before applying to all capsules.

\vspace{-0.3cm}
\begin{align}\tiny \label{eq:n-aff-trans}
\hat{I}_{\text{Cap}} &= W_{\text{Cap}}I_{\text{Cap}} \\
\begin{split} \label{eq:n-tmprry-val}
b\ &=\ b\ +\ \hat{I}_{\text{Cap}}\cdot v
\end{split}
\end{align}
where $I_{\text{Cap}}$ represents the input Capsules, $W_{\text{Cap}}$ is the weight, $\hat{I}_{\text{Cap}}$ represents the predicted capsules, $b$ represents all temporary values, and $v$ represents the activated capsules. Table~\ref{tab:4} shows the related size of each variable.

\begin{figure} [ht]
\centering
\caption{XnODR (Xnorizes the Linear Projection Outside Dynamic Routing), Version 1 of the proposed Fully Connected layer.}
\includegraphics[width=6cm,height=7.5cm,scale=0.5]{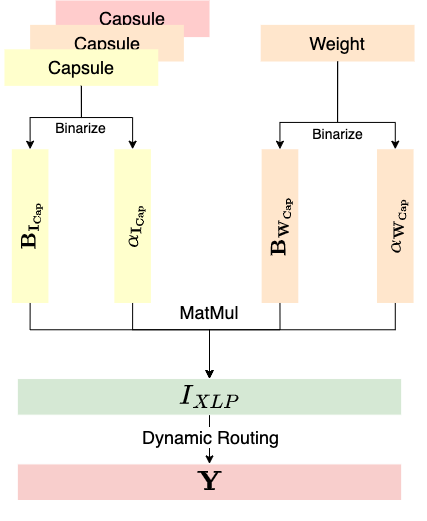}
\label{fig:XnODR}
\vspace{-0.7cm}
\end{figure}

\subsubsection{XnODR(Xnorizes the Linear Projector Outside the Dynamic Routing)} \label{sec:3.3.1} 

The core of XnODR is to xnorize the affine transformation, Eq.~\ref{eq:n-aff-trans}. Let $I_{\text{Prim}}$ denote output tensors from the PrimaryCap layer. Algorithm~\ref{alg:1} summarized the specific steps. $B_{I_{\text{Cap}}}$ and $B_{W_{\text{Cap}}}$ are the binary filters, while $\alpha_{I_{\text{Cap}}}$ and $\alpha_{W_{\text{Cap}}}$ are the scaling factors. We also depict this process in Figure.~\ref{fig:XnODR}

\begin{algorithm}[ht]\small
\caption{XnODR Algorithm}
\begin{algorithmic}[1]
\INPUT $I_{\text{Prim}}$ (3-dimensional)
\OUTPUT $Y$
\State  \textbf{Expand} $I_{\text{Prim}} \rightarrow I_{\text{Cap}}$ (5-dimensional).
\State \textbf{Initialize} $W_{\text{Cap}}$.
\State \textbf{Binarize} $I_{\text{Cap}}$ $\rightarrow$ $B_{I_{\text{Cap}}}$ and $\alpha_{I_{\text{Cap}}}$.
\State \textbf{Binarize} $W_{\text{Cap}}$ $\rightarrow$ $B_{W_{\text{Cap}}}$ and $\alpha_{W_{\text{Cap}}}$. 
\State \textbf{Affine Transformation}:
$I_{\text{Cap}}* W_{\text{Cap}} \approx (B_{I_{\text{Cap}}}\circledast B_{W_{\text{Cap}}})\odot \alpha_{I_{\text{Cap}}}\ \alpha_{W_{\text{Cap}}}$.
\State \textbf{Let} $I_{\text{XLP}}[p,i,j,:,:] = I_{\text{Cap}}* W_{\text{Cap}}$, where $p\in[0,b]$, $i\in[0, \text{caps\_in}]$, $j\in[0, \text{caps\_out}]$. 
\State $Y = \text{Dynamic\_Routing}(I_{\text{XLP}})$.
\end{algorithmic}
\label{alg:1}
\end{algorithm}

Given that we average the last channel of $I_{\text{Cap}}$, the size of $\alpha_{I_{\text{Cap}}}$ is [$bs$, $\text{caps}_{in}$, $\text{caps}_{out}$, $1$, $1$]. When, we average the third channel of $W_{\text{Cap}}$, the size of $\alpha_{W_{\text{Cap}}}$ is [$\text{caps}_{in}$, $\text{caps}_{out}$, $1$, $\text{dim}_{out}$]. In addition, binarization changes the capsule's value instead of its size. Therefore, $B_{I_{\text{Cap}}}$ has the size of [$bs$, $\text{caps}_{in}$, $\text{caps}_{out}$, $1$, $\text{dim}_{in}$], while $B_{W_{\text{Cap}}}$ has the size of [$\text{caps}_{in}$, $\text{caps}_{out}$, $\text{dim}_{in}$, $\text{dim}_{out}$]. $\circledast$ denotes the convolutional operation using XNOR and the bitcount operations. $\odot$ represents the element-wise product. $I_{\text{XLP}}$ represents the results. $\text{Dynamic\_Routing}$ represents the DR. $Y$ represents the final output, where its size is [$bs$, $\text{caps\_in}$, $\text{caps\_out}$, $1$, $\text{dim\_out}$].

Section~\ref{sec:3.1} introduces the detail of the DR mechanism. According to the theory of~\cite{A2_xnor}, the total number of operations in a standard convolution is $cN_{W}N_{I}$, where $c$ is the channel number, $N_{W}=wh,\ N_{I}=w_{in}h_{in}$. "With the current generation of CPUs, we can perform 64 binary operations in one clock of CPU"~\citep{A2_xnor}. The total parameters from the xnorized convolution is $\frac{1}{64}cN_{W}N_{I}+N_{I}$, where $cN_{W}N_{I}$ is the binary operation, $N_{I}$ is the  non-binary operation. The speed-ratio equation, also known as Speed Up, is summarized as Eq.~\ref{eq:spd-rt}. It represents the times of the FLOPs of convolutional operation over the xnorized convolution.

\begin{equation}\small \label{eq:spd-rt}
    S=\frac{cN_{W}N_{I}}{\frac{1}{64}cN_{W}N_{I}+N_{I}}
\end{equation}

In our case, we only compare the operations of the linear projector before DR, since we only binarize this part in \textbf{XnODR}. We take $I_{\text{Cap}}$ as the input for the linear projector, and $N_{I}$ as each capsule's dimension. We also take $W_{\text{Cap}}$ as the related weight, and $N_{W}$ as the product of $\text{dim}_{in}$ and $\text{dim}_{out}$. Therefore, we formulate the operations of the usual linear projector as Eq.~\ref{eq:9}:

\vspace{-0.5cm}
\begin{equation} \label{eq:9}
\text{caps}_{in}\text{caps}_{out}\times \text{dim}_{out}\text{dim}_{in}\text{dim}_{out}.
\end{equation}

The operations of binary one are formulated as Eq.~\ref{eq:10}:
\begin{multline} \label{eq:10}
\frac{1}{64}\text{caps}_{in}\text{caps}_{out}\times \text{dim}_{out}\\ 
\times\text{dim}_{in}\text{dim}_{out}+\text{caps}_{out}.
\end{multline}

Therefore, the speed-ratio is:
\vspace{-0.4cm}
\begin{equation} \label{eq:11}
\frac{\text{caps}_{in}\text{caps}_{out}\text{dim}^{2}_{out}\text{dim}_{in}}{\frac{1}{64}\text{caps}_{in}\text{caps}_{out}\text{dim}^{2}_{out}\text{dim}_{in}+\text{dim}_{in}}.
\end{equation}

We show the result in Section~\ref{sec:4}.

\begin{figure} [ht]
\centering
\caption{XnIDR(Xnorize the Linear Projection Inside Dynamic Routing), the Version 2 of proposed Fully Connected layer.}
\includegraphics[width=5.8cm,height=7.5cm, scale=0.5]{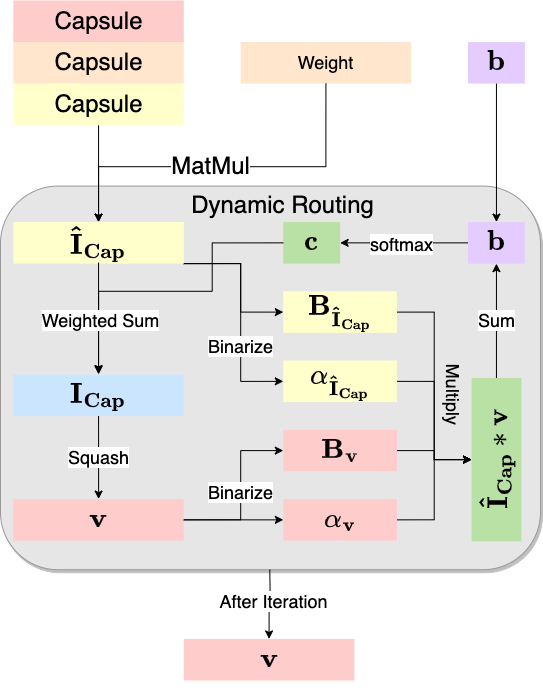}
\vspace{-0.7cm}
\label{fig:XnIDR}
\end{figure}

\subsubsection{XnIDR(Xnorize the Linear Projector Inside Dynamic Routing)}\label{sec:3.3.2}

The core of XnIDR is to xnorize the linear projector inside DR (see Eq.~\ref{eq:n-tmprry-val}). We summarized the whole procedure in Algorithm~\ref{alg:2}. $B_{\hat{I}_{\text{Cap}}}$ and $B_{v}$ are the binary filters, $\alpha_{\hat{I}_{\text{Cap}}}$ and $\alpha_{v}$ are the scaling factors. We plot this algorithm in Figure.~\ref{fig:XnIDR}.

\begin{algorithm*}[ht]\small
\caption{XnIDR Algorithm}
\begin{multicols}{2}
\begin{algorithmic}[1]
\INPUT $I_{\text{Prim}}$ (3-dimensional)
\OUTPUT $v$
\State \textbf{Expand} $I_{\text{Prim}} \rightarrow I_{\text{Cap}}$ (5-dimensional)
\State \textbf{Initialize} $W_{\text{Cap}}$
\State $\hat{I}_{\text{Cap}}=W_{\text{Cap}} I_{\text{Cap}}$
\State \textbf{Initialize} it = 0
\While{it < iteration\_number}
\State $b=0$
\State $c_{ij}\ =\ \frac{exp(b_{ij})}{\sum_{k}exp(b_{ik})}$
\State $I_{Cap_{j}} = \sum_{i}c_{ij}\hat{I}_{Cap_{j|i}}$
\State $v_{j} = \frac{||I_{Cap_{j}}||^{2}}{1+||I_{Cap_{j}}||^{2}}\frac{I_{Cap_{j}}}{||I_{Cap_{j}}||}$
\State \textbf{Binarize} $\hat{I}_{\text{Cap}} \rightarrow B_{\hat{I}_{\text{Cap}}}$ and $\alpha_{\hat{I}_{\text{Cap}}}$
\State \textbf{Binarize} $v \rightarrow B_{v}$ and $\alpha_{v}$ 
\State $b\ =\ b\ +\ B_{\hat{I}_{\text{Cap}}}\circledast B_{v} \odot\alpha_{\hat{I}_{\text{Cap}}}\alpha_{v}$
\EndWhile
\end{algorithmic}
\end{multicols}
\label{alg:2}
\end{algorithm*}

Given that we average the last channel of $\hat{I}_{\text{Cap}}$, the size of $\alpha_{\hat{I}_{\text{Cap}}}$ is [$bs$, $\text{caps}_{in}$, $\text{caps}_{out}$, $1$, $1$]. We average the last channel of $v$, the size of $\alpha_{v}$ is [$bs$, $1$, $\text{caps}_{out}$, $1$, $1$]. In addition, binarization changes the capsule's value instead of size. Therefore, $B_{\hat{I}_{\text{Cap}}}$ has the size of [$bs$, $\text{caps}_{in}$, $\text{caps}_{out}$, $1$, $\text{dim}_{out}$], while $B_{v}$ has the size of [$bs$, $1$, $\text{caps}_{out}$, $1$, $\text{dim}_{out}$]. $v$ represents all the activated capsules, which is the final output.

In the meanwhile, we only compare the operation times of linear projector within DR for Speed Up, since we only xnorize this part in \textbf{XnIDR}. Here, we take $\hat{I}_{\text{Cap}}$ as the input of the linear projector, which is a tensor of capsules, and select $v$ to represent the weight. Therefore, we formulate the operations of the usual linear projector as Eq.~\ref{eq:14}:

\vspace{-0.2cm}
\begin{equation} \label{eq:14}
\text{caps}_{in}\times \text{caps}_{out}\times \text{dim}_{out}^{2}.
\end{equation}

The operations of binary one are formulated as Eq.~\ref{eq:15}:
\begin{equation} \label{eq:15}
\frac{1}{64} \text{caps}_{in} \times \text{caps}_{out} 
\times \text{dim}_{out}^{2} + \text{dim}_{out}.
\end{equation}
\vspace{-0.3cm}

The speed-ratio is formulated as Eq.~\ref{eq:16}:
\begin{equation} \label{eq:16}
    \frac{\text{caps}_{in}\times \text{caps}_{out}\times \text{dim}_{out}^{2}}{\frac{1}{64}\text{caps}_{in}\times \text{caps}_{out}\times \text{dim}_{out}^{2} + \text{dim}_{out}}.
\end{equation}
\vspace{-0.4cm}

The result is shown in Section~\ref{sec:4}.

\subsubsection{Summary} \label{sec:3.3.3}

There are two linear projectors in the original CapsFCLayer \citep{A1_caps}. One is outside DR, while the other is inside DR. XnODR xnorizes the linear projector outside DR. XnIDR xnorizes the one inside DR. Therefore, XnODR and XnIDR are two different variants of CapsFCLayer, which simplify operations by xnorizing linear projectors at different positions. However, XnODR causes more information loss than XnIDR, because XnIDR preserves all information in the outer linear projector. In specific, Eq.~\ref{eq:n-aff-trans} prepares the input values for the following DR in XnODR, which means the information loss may exacerbate during the iterative process. In XnIDR, on the other hand, Eq.~\ref{eq:n-tmprry-val} prepares the temporary value $b$, which is to update $c_{ij}$, which means the information loss has few negative effects on the softmax process directly. Therefore, the information loss of XnIDR has a weaker impact than that of XnODR, which contributes to better performance on accuracy.

Furthermore, we xnorize $\hat{I}_{\text{Cap}}$ (the size is [$bs$, $\text{caps}_{in}$, $\text{caps}_{out}$, $1$, $\text{dim}_{out}$]) and $v$ (the size is [$bs$, $1$, $\text{caps}_{out}$, $1$, $\text{dim}_{out}$]) in XnIDR. In XnODR, we xnorize $I_{\text{Cap}}$ (the size is [$bs$, $\text{caps}_{in}$, $\text{caps}_{out}$, $1$, $\text{dim}_{in}$]) and $W_{\text{Cap}}$ (the size is [$\text{caps}_{in}$, $\text{caps}_{out}$, $\text{dim}_{in}$, $\text{dim}_{out}$]). As we can see, the second dimension and the fourth one of $v$ have 1 channel, while the first dimension and the third dimension of $W_{\text{Cap}}$ has $\text{caps}_{in}$ and $\text{dim}_{in}$ channels. $\text{caps}_{in}$ and $\text{dim}_{in}$ are much larger than one so that XnIDR has less FLOPs during binarization than XnODR. In total, the FLOPs of XnIDR are less than that of XnODR.

Hence, XnIDR is theoretically better than XnODR. Simultaneously, if the network implements Xnorization operation on both linear projectors simultaneously, it predicts awfully due to lacking too much information.

\vspace{-0.2cm}
\section{Experimental Results and Evaluation} \label{sec:4}

In this section, we first introduce the datasets, evaluation metrics, and implementation details. Then, we explain our experiments, present the results, report the ablation study, and analyze the results.

\subsection{Datasets}

We choose three datasets (MNIST, CIFAR-10, and MultiMnist) to evaluate our proposed XnODR and XnIDR. 

\textbf{MNIST}: The National Institute of Standards and Technology was in charge of creating the MNIST dataset~\cite{A50_mnst}. It consists of 70,000 28$\times$28 gray-scale images in 10 classes. There are 60,000 training images and 10,000 test images. The American Census Bureau employees contributed half of the training images, and American high school students contributed the other half. Test images have the same background. The categories are 0, 1, 2, 3, 4, 5, 6, 7, 8, and 9.  

\textbf{CIFAR-10}: This dataset was collected by \cite{A49_cfr10}. It consists of 60,000 32$\times$32 color images in 10 classes, with 6,000 images per class. There are 50,000 training images and 10,000 test images. The categories consist of the airplane, automobile, bird, cat, deer, dog, frog, horse, ship, and truck.

\textbf{MultiMNIST}: This dataset is generated out of MNIST to prove the effectiveness of CapsNet. Our proposed layers get inspired by CapsNet. We, therefore, validate the XnODR and XnIDR by MultiMNIST\footnote{Read code on \hyperlink{https://github.com/jiansfoggy/CODE-SHOW/blob/master/Python/Multi\_Mnist/fast\_generate\_multimnist.py}{Github}}~\citep{A1_caps}.

We create MultiMNIST following the instruction from \cite{A1_caps}, except generating four rather than 1K MultiMNIST examples for each digit in the MNIST dataset, because we find that the model can converge to accuracy higher than 99\% without a large volume dataset. So the training set consists of 240,000 36$\times$36 gray-scale images in 10 classes, and the test set size is 40,000 36$\times$36 gray-scale images in 10 classes. The categories are 0, 1, 2, 3, 4, 5, 6, 7, 8, and 9. MultiMNIST comprises two overlapped digits on each image, while MNIST has one single number per image.

\subsection{Evaluation Metrics} \label{sec:4.2}

We present the prediction accuracy (the mean value and standard deviation among five trained models with different random seeds), the number of network parameters, Speed Up (see Eq.~\ref{eq:spd-rt}), and FLOPs as the evaluation metrics.

\subsection{Implementation Details} \label{sec:4.3}

For the MNIST classification task, we take gray-scale images with the shape of [28, 28, 1] as the input values and convert the labels to categorical values with the size of [$bs$, 10], where $bs$ represents batch size. Likewise, For the CIFAR-10 classification task, we take color images with the shape of [32, 32, 3] as the input values and convert the labels to categorical values with the size of [$bs$, 10]. To enhance the performance, we do random data augmentation on CIFAR-10 before training. For the MultiMNIST classification task, we take gray-scale images with the shape of [36, 36, 1] as the input values and convert labels to categorical values with the size of [$bs$, 10]. 

To better evaluate XnODR/XnIDR, we selected MobileNetV2~\citep{A18_MBV2} and ResNet-50~\citep{A19_RES50} as the backbone models and tested XnODR/XnIDR on them separately. MobileNetV2 is a lightweight model, while ResNet-50 is a well-known heavyweight model. Another goal of these variants is to evaluate XnODR's and XnIDR's effectiveness on different datasets.

The original papers did not test MobileNetV2 and ResNet-50 on MNIST, CIFAR-10, and MultiMNIST. Thereby, we also evaluated MobileNetV2 and ResNet-50 on MNIST, CIFAR-10, and MultiMNIST.

CapsNet uses the \textit{Margin Loss} to help the model converge better, especially in the case of multiple digits. Following their approach, we selected the \textit{Margin Loss} as well. Then, we utilized the Adam Optimizer and cyclic learning rate scheduler. About the scheduler, we set the range of learning rate as [1e-9, 1e-3] with the step size is 6000. The cyclic learning mode is triangular2. The epoch number is 30 for MNIST, 80 for CIFAR-10, and 20 for MultiMNIST.

After training, we collect and record the Top-1 accuracy, trainable parameters, and FLOPs for comparison. Given that MultiMNIST contains two digits per image, we recorded the Top-2 accuracy for it. We coded the network using TensorFlow (2.4.3) framework and ran experiments on the NVIDIA GTX 1080Ti GPU.

\subsection{Experimental Results} \label{sec:4.4}

This section presents and discusses the results of our comprehensive experiments conducted to evaluate our proposed XnODR and XnIDR modules.

\subsubsection{Model Accuracy} \label{sec:4.4.1}

Table~\ref{tab:all-acc} shows that the models with XnODR or XnIDR reached higher accuracy than the original ones. For example, MobileNetV2\_XnIDR achieves 99.74\% mean accuracy with a standard deviation of 0.0002 on MNIST, 0.13\% larger than MobileNetV2 and comparable to HVCs~\citep{A8_hvcs}'s 99.87\%. Moreover, models with XnIDR classifier perform better than those with XnODR. For instance, in CIFAR-10, the mean accuracy of ResNet\_XnIDR (96.32\%) is 0.03\% higher than that of ResNet\_XnODR (96.29\%) and even 2.13\% higher than ResNet-50's 94.19\%. This improvement of 2.13\% prevails that XnIDR has the potential to reach the peak made by "CaiT-M-36 $\gamma$ 224"~\citep{A6_cifar10sota_extra} (99.40\%) once embedding XnIDR into a more advancing backbone. In MultiMNIST, ResNet\_XnIDR ((99.31$\pm$0.03)\%) still ranks first. ResNet\_XnODR ranks No.2 with its result of (99.26$\pm$0.01)\%, followed by ResNet-50, (99.12$\pm$0.03)\%. The high mean value and low standard deviation show the stability of the models with XnODR/XnIDR.

\begin{table*} [ht]
\centering
\caption{shows the accuracy of different models on MNIST, CIFAR10, and MultiMNIST. The results include models from cited papers, and the experiments of baseline and our proposed variants. The display format of baseline and our proposed variants is (mean accuracy $\pm$ standard deviation)\%.}
\begin{tabular}{l|c|c|c}
\toprule
\multirow{2}{*}{Method} & MNIST & CIFAR-10 & MultiMNIST \\
\cmidrule{2-4}
& Top1 & Top1 & Top2 \\
\midrule
Efficient-CapsNet~\citep{A7_efcaps} & 99.84\% & - & 94.90\% \\ 
HVCs~\citep{A8_hvcs} & \textbf{99.87\%} & 89.23\% & - \\
PLM(BCE+CB)~\citep{A12_plm} & - & 70.33\% & 89.37\% \\
CaiT-M-36 $\gamma$ 224~\citep{A6_cifar10sota_extra} & - & \textbf{99.40\%} & - \\
L1/FC~\citep{A13_rtdis_caps} & 99.56\% & 85.96\% & 92.46\% \\
ACGAN~\citep{A59_ACGAN} & 96.25\% & - & - \\
EnsNet~\citep{A9_ensnet} & 99.84\% & 76.25\% & - \\ 
LaNet-L~\citep{A5_cifar10sota} & - & 99.01\% & - \\
SCAE~\citep{A10_scae} & 99.0\% & 33.48\% & - \\
capsnet+PA~\citep{A14_capsPA} & 99.67\% & 85.69\% & 94.88\% \\
CPPN~\citep{A11_cvae} & 97.00\% & - & 65.6\% \\
CapsuleNet~\citep{A1_caps} & 99.65\% & 89.40\% & 94.8\% \\
XNOR-Net~\citep{A2_xnor} & - & 89.83\% & - \\
\midrule
ResNet-50 & (99.57$\pm$0.02)\% & (94.19$\pm$0.1)\% & (99.12$\pm$0.03)\%\\
ResNet\_XnODR & (99.66$\pm$0.02)\% & (96.29$\pm$0.09)\% & (99.26$\pm$0.01)\%\\
ResNet\_XnIDR & (99.67$\pm$0.02)\% & (96.32$\pm$0.05)\% & (\textbf{99.31}$\pm$0.03)\% \\
MobileNetV2 & (99.61$\pm$0.02)\% & (95.39$\pm$0.09)\% & (98.62$\pm$0.05)\% \\
MobileNetV2\_XnODR & (99.73$\pm$0.01)\% & (96.05$\pm$0.04)\% & (99.13$\pm$0.03)\% \\
MobileNetV2\_XnIDR & (99.74$\pm$0.02)\% & (96.14$\pm$0.2)\% & (99.14$\pm$0.02)\% \\
\botrule
\end{tabular} 
\label{tab:all-acc}
\end{table*}

We also compared ResNet\_XnODR/XnIDR with the cited papers in Section~\ref{sec:2} and other variants of ResNet-50 in Table~\ref{tab:resnet-cifar10}. It presents that ResNet\_XnODR/XnIDR perform better than the variants of CapsNet and XNOR-Net cited in Section~\ref{sec:2}. Taking ResNet-50's variants into account, ResNet\_XnIDR ranks No.3 across all listed models, while ResNet\_XnODR reaches fourth place.

\begin{table}[ht]
\caption{shows experiment results on CIFAR-10 from cited papers. * represents the variants of CapsNet and XNOR-Net cited in Section~\ref{sec:2}. ** represents the other ResNet-50 variants.} 
\begin{tabular}{@{}l|c@{}}
\toprule
Models & CIFAR-10 Accuracy \\
\midrule
Aff-CapsNets$^{*}$ \citep{A24_Rob_Caps} & 76.28\% \\
CapsNetSIFT$^{*}$ \citep{A44_CapsnetSIFT} & 91.27\% \\
HGCNet-91$^{*}$ \citep{A43_high_eff} & 94.47\% \\
\multirow{2}{4.cm}{Ternary connect + Quantized backprop$^{*}$ 
 \citep{A37_mul}} & \multirow{2}{*}{87.99\%} \\ 
 & \\
\multirow{2}{4.cm}{Greedy Algorithm for Quantizing$^{*}$ \citep{A39_grdy}} & \multirow{2}{*}{88.88\%} \\
 & \\
SLB on ResNet20$^{*}$ \citep{A40_low_bit} & 92.1\% \\
SLB on VGG small$^{*}$ \citep{A40_low_bit} & 94.1\% \\
DoReFa-Net$^{*}$ on VGG-11 \citep{A41_prune} & 86.30\% \\
DoReFa-Net$^{*}$ on ResNet14 \citep{A41_prune} & 89.84\% \\
\midrule
BootstrapNAS’ ResNet-50~ \citep{A61_BootstrapNAS} & 93.70\% \\
ResNet50(A1)$^{**}$~\citep{A72_ResNet50_A1} & \textbf{98.30\%} \\
ResNet-50-SAM$^{**}$~\citep{A73_ResNet50-Sam} & 97.40\% \\
ResNet-50x1-ACG$^{**}$~\citep{A74_ResNet50x1-ACG} & 95.78\% \\
\midrule
ResNet-50 & 94.19\% \\
ResNet\_XnODR & 96.29\% \\
ResNet\_XnIDR & 96.32\% \\
\botrule
\end{tabular} \label{tab:resnet-cifar10}
\end{table}

\subsubsection{Model FLOPs and PARA} \label{sec:4.4.2}

Table~\ref{tab:all-flops-para} shows that the models with XnODR/XnIDR cost less FLOPs and parameters. For example, ResNet\_XnODR/XnIDR cost lower FLOPs and parameters than ResNet-50 across all three datasets. MobileNetV2\_XnODR/XnIDR also spend less FLOPs and parameters than MobileNetV2 on three datasets. However, MobileNetV2\_XnODR is an exception, it costs more FLOPs than MobileNetV2 on MNIST. To figure out the reason, we computed the FLOPs of FC layers from MobileNetV2 and that of MobileNetV2\_XnODR and recorded all the specific values in Table~\ref{tab:flops-only}.

\begin{table*}[ht]
\centering
\caption{shows the FLOPs and Parameters of ResNet-50, ResNet\_XnODR/XnIDR, MobileNetV2 and MobleNetV2\_XnODR/XnIDR.}
\begin{tabular}{l*{6}{|c}}
\toprule
\multirow{2}{*}{Method} & \multicolumn{2}{c}{MNIST} & \multicolumn{2}{|c}{CIFAR-10} & \multicolumn{2}{|c}{MultiMNIST} \\
\cmidrule{2-7}
& FLOPs & PARA & FLOPs & PARA & FLOPs & PARA \\
\midrule
ResNet-50 & 3.865B & 26.16M & 3.865B & 26.16M & 1.012B & 26.16M \\
ResNet\_XnODR & 3.864B & 23.85M & 3.864B & 23.85M & 1.011B & 23.86M \\
ResNet\_XnIDR & 3.862B & 23.85M & 3.862B & 23.85M & 1.009B & 23.85M \\
MobileNetV2 & 312.25M & 3.05M & 312.25M & 3.05M & 83.62M & 3.05M \\
MobileNetV2\_XnODR & 312.60M & \textbf{2.99M} & 312.60M & \textbf{2.99M} & 83.96M & \textbf{2.99M} \\
MobileNetV2\_XnIDR & \textbf{311.74M} & \textbf{2.99M} & \textbf{311.74M} & \textbf{2.99M} & \textbf{83.11M} & \textbf{2.99M} \\
\botrule
\end{tabular} \label{tab:all-flops-para}
\end{table*}

To be specific, in Table~\ref{tab:flops-only}, we also listed the FLOPs of the binarization process and XNOR operation, which are two parts of Xnorization (Appendices~\ref{appendix:A} and~\ref{appendix:B}). Given that XNOR is a binary operation, we initially calculated its binary point of operations (BOPs) and then divided it by 64 to get the related FLOPs. The ratio to the original FC means that the FLOPs of models with XnODR/XnIDR over that of the original ones. Appendix~\ref{appendix:C} shows the structure of the Fully Connected layers deployed in MobileNetV2 and ResNet-50.

\begin{table*}[ht]
\centering
\caption{shows the FLOPs of the original FC layers and that of XnODR/XnIDR. For the sake of comprehensive comparison, we list the FLOPs of binarization, the BOPs (binary operations) of XNOR operation, the FLOPs of XNOR operation, and the ratio of the FLOPs of XnODR/XnIDR over that of the original FC layers.}
\begin{tabular}{l|c|c|c|c|c} 
\toprule
\multirow{2}{*}{Details} & \multirow{2}{*}{Total} & \multirow{2}{*}{Binarization} & \multirow{2}{*}{XNOR BOPs} & \multirow{2}{*}{XNOR FLOPs} & \multirow{2}{2.cm}{Ratio to original FC} \\
&&&&& \\
\midrule
ResNet-50 & 5.25M & & & & \\
ResNet\_XnODR & 3.97M & 3.13M & 41K & 640 & 75.52\% \\
ResNet\_XnIDR & 2.26M & 811.01K & 82K & 1280 & 43.02\% \\
MobileNetV2 & 1.64M & & & & \\
MobileNetV2\_XnODR & 1.98M & 1.57M & 20K & 320 & 120.89\% \\
MobileNetV2\_XnIDR & 1.13M & 405.50K & 41K & 640 & 68.86\% \\
\botrule
\end{tabular}
\label{tab:flops-only}
\end{table*}

We found that the main reasons are the number of FC layers and the Binarization process. About the number of FC layers, according to Tables~\ref{tab:flops-only} and~\ref{tab:fc_structure}, the current MobileNetV2 has four FC layers that take 1.64M FLOPs. If MobileNetV2 uses the same FC layers as ResNet-50 does, it would have 3.68M FLOPs, which is larger than XnODR's FLOPs, 1.98M. Then, the Binarization process is another reason. Table~\ref{tab:flops-only} shows that XnODR requires 1.98M FLOPs on MobileNetV2\_XnODR, which is 120.89\% of FC layers' FLOPs (1.64M) on MobileNetV2. However, 1.57M out of 1.98M are from binarization. This is the reason that MobileNetV2\_XnODR has more FLOPs than MobileNetV2. Except for binarization, MobileNetV2\_XnODR only requires 20K binary operations (320 FLOPs) of the XNOR operation. This number is too small to affect the total FLOPs of XnODR. 

In general, XnODR/XnIDR helps reduce the FLOPs and parameters on both lightweight models and heavyweight ones.

\subsubsection{Model Speed Up} \label{sec:4.4.3}

Eq.~\ref{eq:spd-rt} is the Speed Up function, which represents the MADD spent by the usual convolutional operation over the XnorConv layer. The larger value means the faster speed. We computed and recorded the Speed Up ratios of models with XnODR/XnIDR in Table~\ref{tab:speed-up}. It shows that the Speed Up of the linear projector in XnIDR is a little less than that of XnODR.

\begin{table}[ht]
\caption{shows speed comparison among proposed models, ResNet-50 with XnODR/XnIDR and MobileNetV2 with XnODR/XnIDR, under different datasets.}
\begin{tabular}{l|c}
\toprule
Models & Speed Up (Unit: ratio) \\
\midrule
ResNet\_XnODR & 63.99 \\
ResNet\_XnIDR & 63.90 \\
MobileNetV2\_XnODR & 63.98 \\
MobileNetV2\_XnIDR & 63.80 \\ 
\botrule
\end{tabular} \label{tab:speed-up}
\end{table}

However, both XnODR and XnIDR have more steps other than linear projection. As Table~\ref{tab:flops-only} shows, XnODR has more FLOPs than XnIDR since it has larger $\text{dim}_{in}$, which causes binarization process costing larger FLOPs than XnIDR does. Hence, models with XnIDR can run faster than those with XnODR.

\subsubsection{Analysis and Evaluation} \label{sec:4.4.4}

Table~\ref{tab:all-acc} supports that models with XnODR/XnIDR outperform the original models. One reason is the affine transformation (Eq.~\ref{eq:aff-trans}, which becomes Eq.~\ref{eq:n-aff-trans} in XnODR). It enriches the features so that the model learns the image from different angles. Eq.~\ref{eq:wght-sum}, on the other hand, transports the needed capsules from the lower level to the higher level. Then, Eq.~\ref{eq:squash} helps to enlarge the disparity among different capsules and activate the target capsules. In the meanwhile, Eq.~\ref{eq:tmprry-val}, which becomes Eq.~\ref{eq:n-tmprry-val} in XnIDR, monitors the similarity between predicted capsules and activated capsules (the final output values). This similarity helps update $b_{ij}$. The iterative structure takes the updated $b_{ij}$ to the next round. Sabour~\etal~\cite{A1_caps} combined these processes in the CapsFC layer, which helps improve the prediction. Therefore, the CapsFC layer enhances accuracy. Moreover, Xnorization helps reduce the parameter and speed up the calculation process. Hence, our proposed XnODR and XnIDR suit both lightweight models and heavyweight models.

Then, Table~\ref{tab:speed-up} presents that the Speed Up stays the same for each variant on the three datasets. Taking XnODR into consideration, its input size [BS, 128, 1, 8] stays the same on all three datasets. Moreover, the structure of XnODR stayed the same. 

In general, Sections~\ref{sec:4.4.1} and~\ref{sec:4.4.2} support the view that the heavyweight models with either XnODR or XnIDR enhance the accuracy of the original model with faster speed, while the lightweight models with XnIDR improve the accuracy of the original model with faster speed. In detail, the lightweight models with XnIDR rather than XnODR improve the accuracy of the original model with less FLOPs. Simultaneously, the core of the classification task is predicting well. MobileNetV2\_XnODR outperformed MobileNetV2 on accuracy and parameters across all three datasets. Without good prediction results, it is in vain to increase the inference speed of the models. Additionally, lightweight models have different numbers of FC layers. For those with complex FC layers, we believe XnODR requires less FLOPs than them. Comprehensively, lightweight models with XnODR outperformed the original model as well.

To be brief, we have two take-away here. The first take-away is that both heavyweight models and lightweight models with XnODR/XnIDR perform better than the original models. The second take-away is that model taking XnIDR as an FC layer is better than models taking XnODR because of higher accuracy and lower FLOPs.

\subsection{Ablation Study} \label{sec:4.5}

To show the effectiveness of XnODR and XnIDR, we perform ablation studies to evaluate the influence of different components on the classification task. 

In XnODR and XnIDR, the hyperparameters are the input capsule number, and output capsule number, which are subject to the size of input images, and the number of classes in the corresponding datasets. About the capsule dimension and routing number, we followed~\cite{A1_caps} to set them for a fair comparison, as our goal is to show the efficiency of XnODR/XnIDR with the minimum change.

\textbf{Influence of Dynamic Routing}: Dynamic Routing mechanism is the basic framework of XnODR and XnIDR. To show the necessity of proposing XnODR and XnIDR, we do the experiment by replacing the regular FC layers from ResNet-50 and MobileNetV2 with DR.

\begin{table}[ht]
\caption{shows experiment results on models with the typical DR mechanism.}
\begin{tabular}{@{}l*{3}{|c}@{}}
\toprule
\multirow{2}{*}{Method} & \multicolumn{3}{c}{MNIST} \\
\cmidrule{2-4}
& Top1 & FLOPs & PARA \\
\midrule
ResNet\_DR & (\textbf{99.69$\pm$0.05})\% & 3.861B & 23.86M \\
MobileNetV2\_DR & (99.57\%$\pm$0.03)\% & 311.29M & 2.99M \\
\midrule
\multirow{2}{*}{Method} & \multicolumn{3}{c}{CIFAR-10} \\
\cmidrule{2-4}
& Top1 & FLOPs & PARA \\
\midrule
ResNet\_DR & (\textbf{96.19$\pm$0.07})\% & 3.861B & 23.86M \\
MobileNetV2\_DR & (96.04$\pm$0.1)\% & 311.29M & 2.99M \\
\midrule
\multirow{2}{*}{Method} & \multicolumn{3}{c}{MultiMNIST} \\
\cmidrule{2-4}
& Top1 & FLOPs & PARA \\
\midrule
ResNet\_DR & (\textbf{99.27$\pm$0.009})\% & 1.009B & 23.86M \\
MobileNetV2\_DR & (99.15$\pm$0.04)\% & 82.66M & 2.99M \\
\botrule
\end{tabular} \label{tab:Abl-DR}%
\end{table}

Comparing Table~\ref{tab:all-acc} with Table~\ref{tab:Abl-DR} shows that models with XnODR/XnIDR generally perform better than those with DR. In addition, it is hard to say models with DR entirely surpass the original ones. For example, models with DR were better than the original models in five related sub-experiments. However, in Table~\ref{tab:Abl-FLOPs}, we noticed that models with DR cost fewer FLOPs than those with XnODR/XnIDR. It is the binarization process in XnODR/XnIDR rather than the XNOR operation requiring too many FLOPs. For example, Table~\ref{tab:flops-only} shows that the FLOPs of XnODR in ResNet\_XnODR is 3.97M, of which 3.13M is from binarization. The remainder is merely 0.84M, which is much less than ResNet\_DR's 1.37M. Moreover, blindly pursuing speed damages our core target of good prediction, which means a high accuracy.

\begin{table}[ht]\small
\caption{compares the FLOPs of ResNet-50 and ResNet\_DR. It also compares the FLOPs of MobileNetV2 and MobileNetV2\_DR.}
\begin{tabular}{l|c}
\toprule
Datasets & FLOPS \\
\midrule
ResNet-50  & 5.25M \\
ResNet\_DR & 1.37M  \\
MobileNetV2 & 1.64M  \\
MobileNetV2\_DR & 682.9K \\ 
\botrule
\end{tabular} \label{tab:Abl-FLOPs}
\end{table}

In summary, models with XnODR/XnIDR are better than those with DR. It is useful to utilize XnODR/XnIDR as new FC layers.

\section{Discussion} \label{sec:5}

The CNNs with our proposed FC layers (XnODR and XnIDR) demonstrate that they outperform the original models both in terms of accuracy and speed by their solid performance in our experiments. For example, ResNet\_XnODR/XnIDR achieves over 96.3\% accuracy with the cost of 3.864B and 3.862B FLOPs and 23.86M parameters on CIFAR-10 while ResNet-50 returns accuracy lower than 96\% with more FLOPs and parameters. As we can see, fusing xnorization into the DR mechanism helps speed the model while maintaining a comparable or even better accuracy. Hence, we can implement XnODR/XnIDR as effective FC layers in both lightweight and heavyweight models. 

XnODR/XnIDR can do more than we introduced above. In order to improve the network's representative capability, either XnODR or XnIDR can work as a parallel branch in CNNs to provide rotation invariance, increase accuracy and avoid loss of time. In addition, it is helpful to load its output into the relabeling mechanism as a contrast.

Both CapsNet and XNOR-net have shown relatively poor performance on large-scale datasets such as ImageNet and CIFAR-100 \cite{A66_CapsNet_CIFAR100}. Similarly, our proposed XnODR/XnIDR layers share the same drawback. Specifically, when using the Xnorization method, the network loses too much information ~\citep{A2_xnor, A17_Xnor-Prune} and that would affect the model capacity, which eventually prevents one from achieving high performance on such datasets.

\section{Conclusion and Future Work} \label{sec:6}

High accuracy and fast processing speed are two highlights of CapsNet and XNOR-Net. Combining these two advantages helps to speed up training while maintaining good or even better performance. Inspired by this idea, we proposed XnODR/XnIDR as the alternative option for the usual FC layer. Then, we inserted them into the MobileNetV2 and ResNet-50 and experimented on three datasets, MNIST, CIFAR-10, and MultiMnist. The results show that the models with XnODR/XnIDR predict better than the original model by taking fewer parameters and less FLOPs. 

In the future, we would work on creating a new Xnorization algorithm to approximate convolutional operation and avoid overt information loss, while simultaneously, we fuse this new Xnorization method and EM Routing \citep{A3_emr} to create a new xnorized CapsConv layer. Finally, we plan to build a new network with the XnorCapsConv layer and XnODR/XnIDR. The target is to achieve good performance on large-scale complex datasets such as AffectNet \citep{A52_affnet} and ImageNet~\citep{A48_img_net}.

\backmatter

\bmhead{Acknowledgments}

We would like to thank Ms. Druselle May who helped us proofread the manuscript.

\section*{Statements and Declarations}

\textbf{Authors' contributions} Jian Sun: Conceptualization, Methodology, Software, Validation, Formal analysis, Investigation, Data Curation, Writing - Original Draft, Writing - Review \& Editing, Visualization.

Ali Pourramezan Fard: Writing - Review \& Editing.

Mohammad H. Mahoor: Resources, Writing - Review \& Editing, Supervision, Project administration.

\textbf{Funding} Not applicable

\textbf{Conflict of interest} The authors have no relevant financial or non-financial interests to disclose.

\textbf{Ethics approval} Not applicable

\textbf{Consent to participate} Not applicable

\textbf{Consent for publication} Not applicable

\begin{appendices}

\section{Binarization} \label{appendix:A}

Xnorization speeds the calculation by three steps. The first step is to binarize the input values and weights before the convolutional operation. Secondly, it introduces a binary dot product with XNOR-Bitcounting operations. The last step is to replace the multiplication operations with the addition operation during the convolutional operation. Simply stated, it has two parts, the binarization process and the XNOR operation. We introduce binarization first and include the XNOR operation in Appendix~\ref{appendix:B}.

The binarization process is to split the tensor into 2 parts. One is sign matrix (spanned from 2 values \{-1, 1\}), and the other one is a scaling factor. 

Let $\mathcal{I}$ be a set of tensors. And $\mathbf{I}=\mathcal{I}_{l(l=1,...,L)}, \mathbf{I}\in\mathbb{R}^{c\times w_{in}\times h_{in}}$ represents the input tensor for the $l^{th}$ layer of network, where $(c, w_{in}, h_{in})$ means \textit{channel}, \textit{width} and \textit{height}. We split the tensor $\mathbf{I}$ into two values, binary filter $\mathbf{B}\in\{+1, -1\}^{c\times w_{in}\times h_{in}}$ and scaling factor $\alpha \in \mathbb{R}^{+}$, and use them to estimate $\mathbf{I}\approx \alpha \mathbf{B}$. 

We first discuss the Sign and the binary filter. According to \citep{A2_xnor} $k$-\textbf{bit Quantization} is $q_{k}(x)=2(\frac{[(2^{k}-1)(\frac{x+1}{2})]}{2^{k}-1} - \frac{1}{2})$. The sign function is 1-bit Quantization, such that $q_{1}(x)=2(\frac{[(2^{1}-1)(\frac{x+1}{2})]}{2^{1}-1} - \frac{1}{2})=2(\frac{x+1}{2}-\frac{1}{2})$, where the inner function, $\frac{x+1}{2}$, is Hard Sigmoid function, the outer function, $2(Y-\frac{1}{2})$, is Tanh function. Therefore, the sign function can be formulated as shown in Eq.~\ref{eq:17}
\vspace{-0.2cm}
\begin{equation} \label{eq:17}
\begin{split}
\mathbf{B}_{\mathbf{HS}} &= Hard\_Sigmoid(Round(\mathbf{I_{Norm}})) \\
 &= \frac{Round(\mathbf{I_{Norm}}) + 1}{2}
\end{split}
\end{equation}
where $\mathbf{B_{HS}}$ is the output of Hard Sigmoid, $\mathbf{I_{Norm}}$ is the Min-Max Normalization result of $\mathbf{I}$. Its range is [0,1]. The round function will round a value bigger than 0.5 to be 1, less than or equal to 0.5 to be 0. And it leaves $\mathbf{I_{Norm}}$ only 2 values, 0 and 1 after rounding, then the output of Hard Sigmoid function $\mathbf{B}_{\mathbf{HS}}\in\{0.5,1\}$. To control the value of $\mathbf{B}_{\mathbf{HS}}$ between 0 and 1, we call the Clip function and round its output, such that
\vspace{-0.2cm}
\begin{align} \label{eq:18}
\mathbf{B}_{\mathbf{C}} &= Clip(\mathbf{B}_{\mathbf{HS}}) = max(0, min(1, \mathbf{B}_{\mathbf{HS}})) \\
\mathbf{B}_{\mathbf{R}} &= Round(\mathbf{B}_{\mathbf{C}}).
\end{align}
\vspace{-0.3cm}

Therefore, we get $\mathbf{B}_{\mathbf{R}}$, which only has 2 values, 0 and 1. To get the expected binary filter $\mathbf{B}$, we load $\mathbf{B}_{\mathbf{R}}$ into Tanh function, $\mathbf{B} = Tanh(\mathbf{B}_{\mathbf{R}}) = 2\times \mathbf{B}_{\mathbf{C}} -1\in\{-1,+1\}$. Now, we calculate the sign of $\mathbf{I}$ out.

About the scaling factor, according to \cite{A2_xnor}, we use the average of $\mathbf{I}$ to represent it.
\vspace{-0.2cm}
\begin{equation} \label{eq:19}
\begin{split}
\alpha &= \frac{1}{n}(\mathbf{I}^{\mathbf{T}}\mathbf{B}) = \frac{\sum |\mathbf{I}_{i}|}{n} = \frac{1}{n}||\mathbf{I}||_{L_{1}}\ \ (L1-Norm)
\end{split}
\end{equation}
\vspace{-0.3cm}

Eq.~\ref{eq:19} is the formula to get the scaling factor, where $\alpha$ represents the scaling factor. 

\section{XnorConvLayer}
\label{appendix:B}

XnorConvLayer is similar to the standard Conv layer, except binaring input and weight before doing convolution. Moreover, we use the XNOR operation to do convolution in the XnorConvLayer. We formulate it as the following.

Let $\mathbf{I}_{j}$ denote $j^{th}$ tensor of $\mathbf{I}$, $\alpha_{\mathbf{I}_{j}}$ denote $j^{th}$ scaling factor, $\mathbf{B_{I}}$ denote binary filter of $\mathbf{I}$. Then $\mathbf{I}\approx A_{\mathbf{I}}\mathbf{B_{I}}$ is the estimate of $\mathbf{I}$ after xnorize, where $A_{\mathbf{I}}=\{ \alpha_{\mathbf{I}_{0}}, \alpha_{\mathbf{I}_{1}},..., \alpha_{\mathbf{I}_{h_{in}}}\}$. 

Then, let $\mathcal{W}$ be a set of tensors, and $\mathbf{W}$ represent the $k^{th}$ weight filter in the $l^{th}$ layer of the network such that $\mathbf{W}=\mathcal{W}_{lk(k=1,...,K^{l})}$. $K^{l}$ is the number of weight filters in the $l^{th}$ layer of the network. What's more, $\mathbf{W}\in\mathbb{R}^{c\times w\times h}$, where $w\leq w_{in}, h\leq h_{in}$. 

Next, we start estimating $\mathbf{W}$ with binary filter, $\mathbf{B_{W}}$, and scaling filter, $A_{\mathbf{W}}$, such that $\mathbf{W}\approx A_{\mathbf{W}}\mathbf{B_{W}}$.

$A_{\mathbf{W}}=\{ \alpha_{\mathbf{W}_{0}}, \alpha_{\mathbf{W}_{1}},..., \alpha_{\mathbf{W}_{j}}, ..., \alpha_{\mathbf{W}_{h_{in}}}\}$, where $\alpha_{\mathbf{W}_{j}}$ denote $j^{th}$ scaling factor.

According to XNOR-Net \citep{A2_xnor}, Xnorization replaces multiplication in convolutional operations with additions and subtractions. And it causes 58× faster convolutional operations and 32× memory savings. This process is called \textbf{Binary Dot Product}.

To approximate the dot product between $\mathbf{X_{1}}$ and $\mathbf{X_{2}}$, such that $\mathbf{X_{1}}^{T}\mathbf{X_{2}}\approx \alpha_{1}\mathbf{B_{1}}^{T}\alpha_{2}\mathbf{B_{2}}$, where $\mathbf{B_{1}},\mathbf{B_{2}}\in$ $\{+1, -1\}^{n}$, $\alpha_{1},\ \alpha_{2}\in \mathbb{R}^{+}$, the paper solved and proved the following optimization:
\vspace{-0.2cm}
\begin{equation} \label{eq:20}
\alpha_{1}^{*},\mathbf{B_{1}^{*}},\alpha_{2}^{*},\mathbf{B_{2}^{*}}= \underset{\alpha_{1},\mathbf{B_{1}},\alpha_{2},\mathbf{B_{2}}}{argmin}||\mathbf{X_{1}}\odot\mathbf{X_{2}} - \alpha_{1}\alpha_{2}\mathbf{B_{1}}\odot\mathbf{B_{2}}||
\end{equation}
where $\odot$ represents the element-wise product.

In the meanwhile, for input tensors, $\mathbf{I}$, and weight, $\mathbf{W}$, we need to compute scaling factor, $\alpha_{I_{j}}$, for all possible sub-tensors in $\mathbf{I}$ with the same size as $\mathbf{W}$ during convolution. To overcome the redundant computations caused by overlaps between sub-tensors, the paper firstly computed a matrix $\mathbf{M_{I}}=\frac{\sum|\mathbf{I}_{:,:,i}|}{c}$, which is the average over absolute values of the elements in the input $\mathbf{I}$ across the channel, c. Then the paper convolved $\mathbf{M_{I}}$ with a 2D filter $\mathbf{k}\in \mathbb{R}^{w\times h}$, $\mathbf{A_{I}} = \mathbf{M_{I}}*\mathbf{k}$, where $\forall ij\ \mathbf{k_{ij}}=\frac{1}{w\times h}$ and $*$ is a convolutional operation. $\mathbf{A_{I}}$ contains scaling factors $\alpha_{\mathbf{I}_{j}}$ for all sub-tensors in the input I.

The above description proves that it makes sense to estimate $\mathbf{I}*\mathbf{W}$ by $(\mathbf{B_{I}}\ \circledast\ \mathbf{B_{W}}) \odot A_{\mathbf{I}}\alpha_{\mathbf{W}}$, which can be formulated as Eq.~\ref{eq:21}:
\vspace{-0.2cm}
\begin{equation} \label{eq:21}
\mathbf{I}*\mathbf{W} \approx (\mathbf{B_{I}}\ \circledast\ \mathbf{B_{W}}) \odot A_{\mathbf{I}}\alpha_{\mathbf{W}}
\end{equation}
where $\circledast$ denotes the convolutional operation using XNOR and the bitcount operations. 

\section{Structure of Fully Connected layers on different models}
\label{appendix:C}

The structure of Fully Connected layers on the typical ResNet-50 and MobileNetV2 deeply affects the paper's conclusion. Hence, we list the detail in Table~\ref{tab:fc_structure}.

\begin{table}[ht]
\caption{shows the structure of all FC layers on ResNet-50 and MobileNetV2}\label{tab:fc_structure}
\begin{tabular}{@{}l|c|c@{}}
\toprule
Models & ResNet-50 & MobileNetV2 \\
\midrule
FC1 & 1024 & 512 \\
FC2 & 512 & 256 \\
FC3 & 10 & 128 \\
FC4 & - & 10 \\ 
\botrule
\end{tabular}
\end{table}




\end{appendices}


\bibliography{refs}% common bib file

%{figures/JS_Photo} 
\begin{wrapfigure}{l}{20mm} 
\includegraphics[width=1.in,height=1.25in,clip,keepaspectratio]{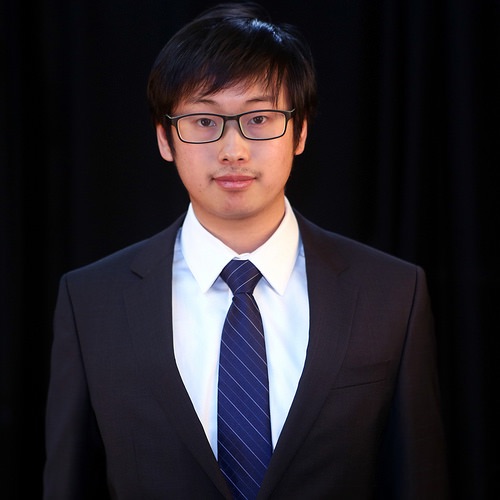}
\end{wrapfigure}\par
\textbf{Jian Sun} received a B.S. in Mathematics and Applied Mathematics from Shandong Agricultural University in 2014, and received an M.S. in Statistics from the George Washington University in 2017. He is currently pursuing a doctoral degree in Computer Science at the University of Denver. His research interests include Computer Vision, Facial Expression Recognition, Facial Landmark Points Detection, Model Framework Design, and Deep Learning.

\begin{wrapfigure}{l}{16mm} 
\includegraphics[width=1.5in,height=1.25in,clip,keepaspectratio]{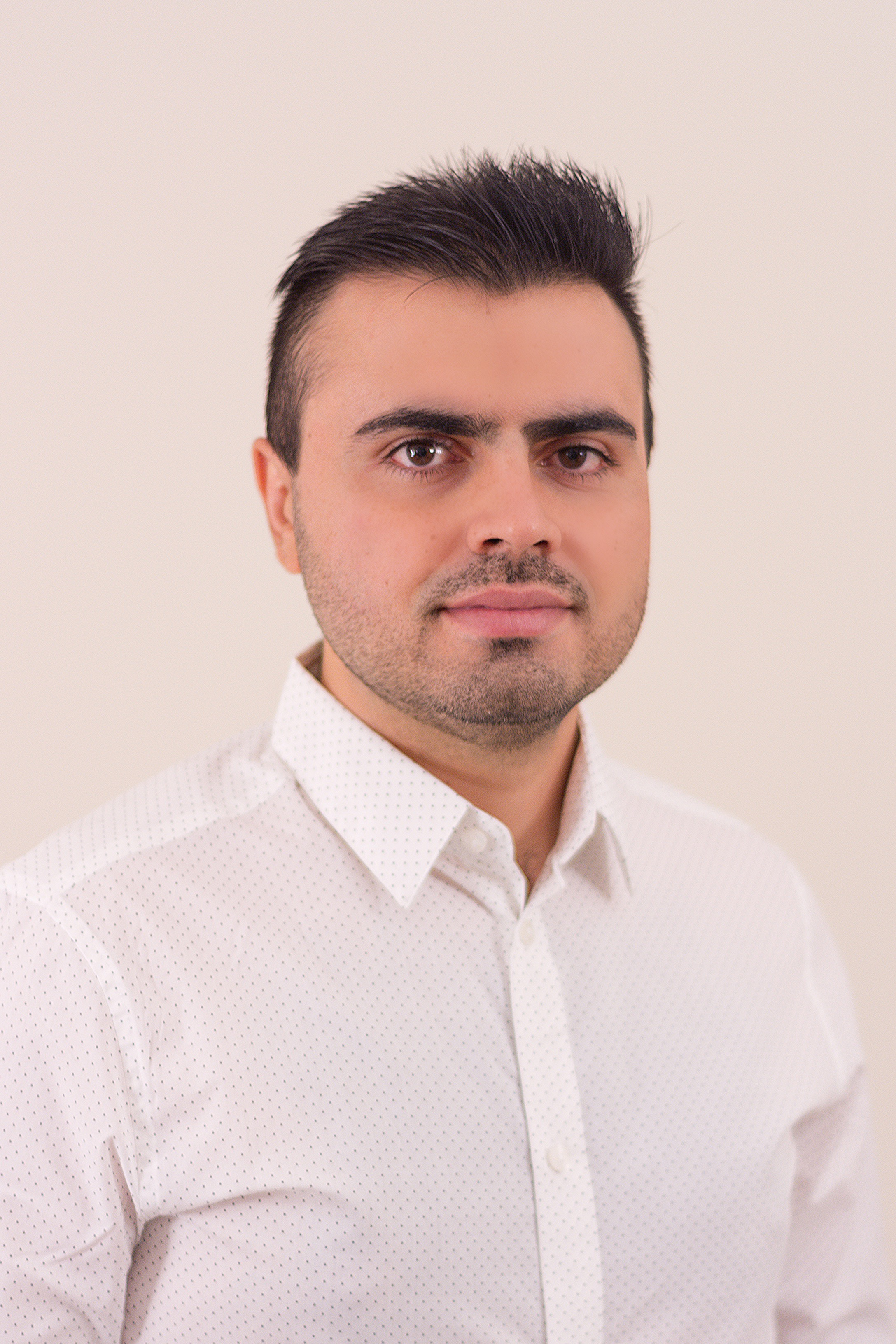}
\end{wrapfigure}\par
\textbf{Ali Pourramezan Fard} received the MSc degree in Computer Engineering - from the Iran University of Science and Technology, Tehran, Iran, in 2015. He is currently pursuing his Ph.D. degree in Electrical \& Computer engineering and is a graduate teaching assistant in the Department of Electrical and Computer Engineering at the University of Denver. His research interests include Computer Vision, Machine Learning, and Deep Neural Networks, especially in face alignment, and facial expression analysis.

\begin{wrapfigure}{l}{20mm} 
\includegraphics[width=1.25in,height=1.3in,clip,keepaspectratio]{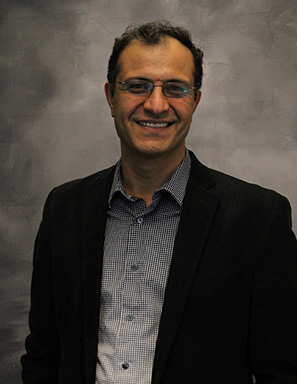}
\end{wrapfigure}\par
\textbf{Mohammad~H.~Mahoor} (Senior Member, IEEE) received the M.S. degree in biomedical engineering from the Sharif University of Technology, Iran, in 1998, and the Ph.D. degree in electrical and computer engineering from the University of Miami, Florida, in 2007. Currently, he is a Professor in electrical and computer engineering with the University of Denver. He does research in the area of computer vision and machine learning, including visual object recognition, object tracking, affective computing, and human–robot interaction (HRI), such as humanoid social robots for interaction and intervention of children with autism and older adults with depression and dementia. He has received over \$7M in research funding from state and federal agencies, including the National Science Foundation and the National Institute of Health. He has published over 158 conferences and journal articles.

\end{document}